\documentclass[10pt,twocolumn,letterpaper]{article}
\usepackage[utf8]{inputenc}
\usepackage[T1]{fontenc}
\usepackage{mathptmx}
\usepackage[margin=0.67in,columnsep=0.25in]{geometry}
\usepackage{graphicx,booktabs,amsmath,amssymb}
\usepackage[numbers,sort&compress]{natbib}
\usepackage{xurl}
\usepackage[hidelinks]{hyperref}
\usepackage{microtype}
\usepackage{enumitem}
\setlist{nosep,leftmargin=1.2em}
\newcommand{\ci}[2]{[\ensuremath{#1},\,\ensuremath{#2}]}
\title{From Probabilities to Decisions:\\Search and Multi-Teacher Distillation with Jev}
\author{Mohamad Yazan Sadoun \quad Sarah Sharif \quad Yaser Mike Banad\\
INQUIRE Lab, School of Electrical and Computer Engineering\\
University of Oklahoma, Norman, OK, USA}
\date{}
\begin{document}
\maketitle

\begin{abstract}
Probability-only models, which TypeSafe calls System One models, return calibrated probabilities for fixed choices in milliseconds and generate no text.
We study one such model, Jev, through two tasks that require decisions under tight constraints.
In bullet chess, a bot that places Jev's judgment inside Stockfish search alongside an opening book and endgame tablebases climbs above a 2200 Lichess bullet rating against other bots.
Live model calls are too slow for search, so we distill pairwise judgments into a compact evaluator that runs at every position.
We then ask how best to spend a fixed labeling budget when an LLM, Qwen3-32B, is available as a second teacher.
In chess, averaging both judges' labels beats spending the whole budget on Qwen alone by 9.6 Elo (95\% interval 4.3 to 14.9), and the gain replicates on fresh openings; a second answer from the same judge is no substitute, and Jev is the strongest partner for Qwen among the models tested.
In passage reranking, Jev's labels alone train a reranker that scores as high as Qwen's, from 21 minutes of API calls instead of 5.1 GPU-hours, and adding Qwen gains at most a few thousandths in ranking quality.
Search supplies the lookahead, distillation makes the judgment cheap enough to use at every position, and an LLM partner pays off in chess.
\end{abstract}

\section{Introduction}

A system that acts under time pressure needs to compare alternatives, anticipate their consequences, and choose before its time runs out.
A probability-only model offers one part of that process: a judgment about specified choices.
TypeSafe's System One model, Jev, returns probabilities without generating text~\cite{typesafe2026docs}.
The name echoes the fast, intuitive System 1 of dual-process accounts of human thinking~\cite{kahneman2011thinking}.
Jev's interface raises a direct question: how much can a decision system do with these judgments?

We investigate that question by connecting Jev to tools that supply what a fixed-choice answer leaves to the surrounding system.
Game-tree search explores future consequences, stored opening and endgame knowledge handles familiar situations, and a learned evaluator scores the positions search reaches.
Bullet chess is a demanding proof of concept because long-term consequences must be considered within a short clock.
The reasoning process we study is this composition of search, knowledge, and evaluation.

The practical obstacle is repeated access to the model.
Even a fast remote judgment is too expensive to request at every position in a search tree.
We therefore collect preferences offline and distill them into a compact evaluator.
This also gives us a place to combine Jev with an LLM: average their probabilities before fitting the evaluator, then use the evaluator throughout search.
The question becomes whether the second judge improves the decisions made by the complete system for the same labeling budget.

The paper follows this progression from a deployed bot to controlled tests of the learned evaluator and its LLM partnership.
The chess experiments show a benefit from combining Jev with Qwen, including against a single judge allowed twice as many training pairs.
Passage reranking tests the broader use of probability-only labels with a different model and task.
There, Jev alone trains a useful reranker, and adding Qwen gains at most $+0.004$.
Together, the two tasks show that distilled probability-only judgments become reusable evaluators, and that an LLM partner's value depends on the task.

\section{Putting Jev's Judgment to Work}
\label{sec:system}

\subsection{Chess as a proof of concept}

Our initial bot connects Jev-derived evaluation to Stockfish search~\cite{stockfish}.
Search follows candidate moves and replies; material values account for the pieces on the board, and a 411-weight evaluator trained from Jev supplies positional preferences.
A masters opening book supplies established continuations, while Syzygy tablebases supply exact information in covered endgames~\cite{syzygy}.
The bot also queries Jev for choices among near-equal moves.
These components put model judgment inside a system that can look ahead and use stored chess knowledge.

\begin{figure}[t]
\centering
\includegraphics[width=\columnwidth]{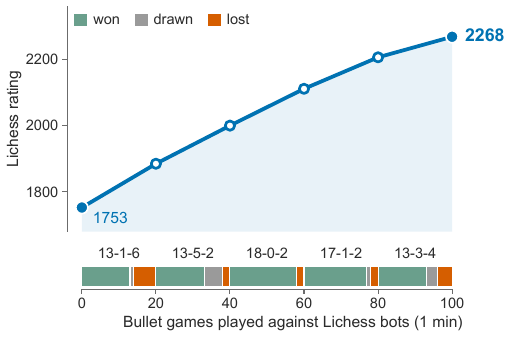}
\caption{The complete bot's Lichess bullet trajectory at one minute per side. The strip groups wins (green), draws (grey), and losses (orange) by deployment round.}
\label{fig:lichess}
\end{figure}

The bot's recorded Lichess bullet trajectory rises from 1753 to 2268 (Figure~\ref{fig:lichess}; Appendix~\ref{app:live}).
This is a result for the complete bot, with all of those components available.
To measure what learned judgment contributes, the subsequent experiments use a common search engine and disable the opening book, tablebases, and live queries during play.
They also use the expanded 443-weight evaluator defined below.

\subsection{Distilling judgment for fast search}
\label{sec:query}

The deployment exposes a mismatch between model access and search.
A live Jev call takes 177\,ms at the median, while search evaluates thousands of positions per move.
In a controlled test, even querying Jev only to break exact ties among the search's best moves reduces strength by 28.7 Elo (normal 95\% interval \ci{-51.1}{-6.6}) over 400 games and 3,875 calls.
A useful judgment must reach the search at a cost the clock can support.

Distillation moves that cost to training.
We ask the model to compare moves offline, fit a small evaluator to its probabilities, and store the resulting weights in the engine.
Search can then use what the evaluator learned at every evaluated position without waiting for a model response.
The same training step can absorb an LLM's judgments, so a partnership need not add any model calls during play.

\section{Distilling One or More Judges}
\label{sec:method}

\paragraph{A shared judgment interface.}
Let $s$ be a position and $a,b$ two candidate moves.
Each judge answers ``Which of these two moves is better for the side to move?'' in both presentation orders; we map the answers to $p(a\succ b)$ and average them.
Jev supplies the probabilities directly for the specified choices~\cite{typesafe2026docs}.
Qwen3-32B~\cite{qwen3card} receives the same board text and question, with reasoning mode off; we normalize the probabilities it assigns to the two move answers.
OLMo-2-0325-32B-Instruct~\cite{olmo2card} follows the same procedure.
DiffusionGemma-26B~\cite{diffusiongemmacard} runs through djev~\cite{djev}, a service that returns probabilities for specified choices.

\paragraph{Combining labels.}
For the Jev--Qwen partnership, the training target is the mean
\begin{equation}
p_F(a,b)=\tfrac12\bigl(p_J(a,b)+p_Q(a,b)\bigr).
\end{equation}
Single-judge evaluators use that judge's probabilities; other partnerships use the corresponding average, a form of multi-teacher distillation~\cite{hinton2015distilling,you2017multiteacher}.
This combination isolates whether a second judge's answers improve the evaluator.

\paragraph{A compact evaluator.}
Let $x(s,a)$ describe the position after move $a$, from the mover's perspective.
Writing $x(s)$ for the features of position $s$, the evaluator computes a weighted sum $t_w(s)=w^\top x(s)$, fitted by
\begin{equation}
\min_w \sum_{(s,a,b)} \ell\!\left(p(a,b),\,\sigma(w^\top\Delta x)\right)+\frac{\lambda}{2}\lVert w\rVert_2^2,
\label{eq:fit}
\end{equation}
where $\Delta x=x(s,a)-x(s,b)$ and $\ell$ is binary cross-entropy against the judge's probability.
This is a Bradley--Terry preference model~\cite{bradley1952rank}; $\lambda$ controls the penalty on weight size.
The evaluator has 443 weights: 384 piece-square weights and 59 for mobility, pawn structure, king attacks, threats, and other positional features.
We centre each piece's square weights at zero to preserve fixed material values.

\begin{figure*}[t]
\centering
\includegraphics[width=\textwidth]{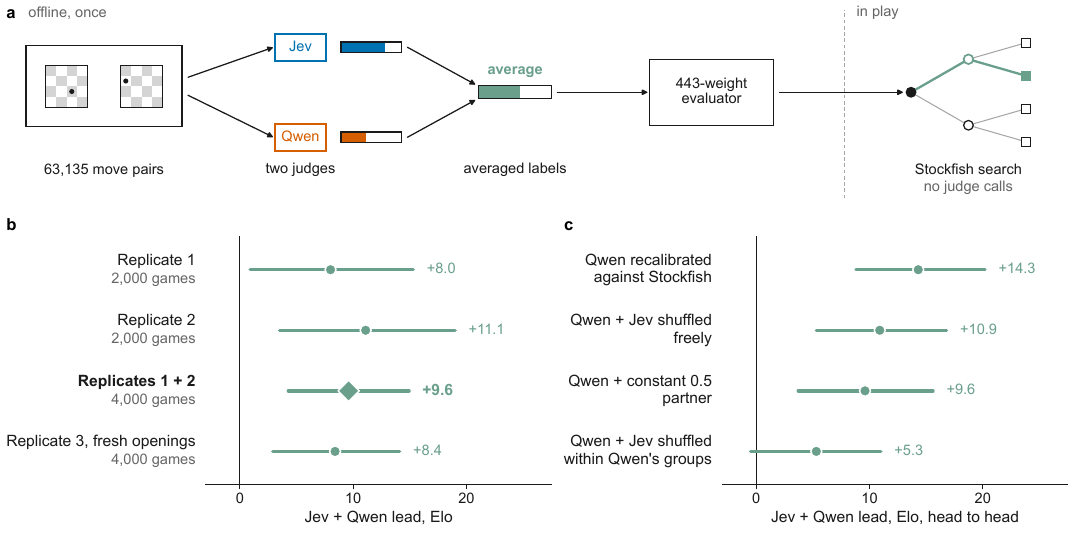}
\caption{(a)~Distilling judge probabilities into search weights. (b)~Combined evaluator vs.\ Qwen on twice the pairs at equal budget; 2,000, 2,000, 4,000 (pooled), and 4,000 games. (c)~Substitute controls, 4,000 games each. Dots: estimates; lines: 95\% intervals.}
\label{fig:headline}
\end{figure*}

\paragraph{Using the evaluator in search.}
The engine evaluates
\begin{equation}
E(s)=m(s)+H(s)+\alpha\,t_w(s),
\label{eq:engine}
\end{equation}
where $m$ scores material, $H$ is a hand-written positional baseline, and $\alpha$ sets the learned score's scale.
Before play, we fold the baseline and learned terms into stored integer weights.
Stockfish 17.1 uses this evaluator in place of its neural-network evaluation; all evaluators run in the same executable with their own weight files.
Correction history is disabled so that leaf scores come from the evaluator alone.
The deployed evaluator requires neither judge (Figure~\ref{fig:headline}a).

\section{Controlled Evaluation}
\label{sec:protocol}

The controlled study changes the learned judgment while holding the search and evaluator form fixed.
It measures whether a label source improves play, whether a partnership is worth its additional answers, and whether that result survives changes to the labels or partner.

\paragraph{Training pairs and response budgets.}
Four judges label 63,135 candidate pairs drawn from 13,000 positions in 2,042 source games.
Candidates are the up-to-four best moves of a material-only Stockfish search (MultiPV~4, 0.1\,s per position, one thread) within 20 centipawns of its top line.
We ask every unordered pair in both presentation orders.
The prompt lists occupied squares, the side to move, and the two moves (Appendix~\ref{app:prompt}).
Each judge supplies 126,270 probabilities, so Jev+Qwen uses 252,540 responses.
For an equal-budget comparison, Jev and Qwen each label a second block of 63,135 pairs on new positions.
Its evaluator then receives the same number of responses as the combined evaluator.

\paragraph{Fitting and baselines.}
We split by source game, holding out one fifth for evaluation and using a further split of the rest to select $\lambda$ by validation log loss.
Every judge uses the same code, features, split rule, and candidate penalties (Appendix~\ref{app:fit}).
The weaker baseline, $H_m$, counts knight, bishop, rook, and queen moves, each weighted 1.0.
The stronger baseline, $H_s$, adds the Simplified Evaluation Function's piece-square values and material adjustments~\cite{simplifiedeval}.
It beats $H_m$ by 55.2 Elo \ci{29.6}{81.4}.
Each comparison holds the baseline fixed and changes the learned addition.

\paragraph{Scale selection.}
The learned score must work at the scale expected by the search.
Adding it to $H_s$ at full scale reduces playing strength; rescaling $H_s$ alone costs $-49.0$ Elo \ci{-72.2}{-26.1}.
We set each learned score's standard deviation to $c\in\{0.125,0.25,0.5,1\}$ times that of $H_s$, play 200 validation games per setting against $H_s$, and select the best, preferring the smaller value in a tie.
All evaluators receive the same calibration budget and validation openings.

\paragraph{Playing strength.}
Games use 60 seconds per side, no increment, one thread, and a 16\,MB position cache; each engine manages its own time.
We play every opening with both colours and count draws as half a point.
For a fraction $p$ of points scored, the Elo difference is $400\log_{10}(p/(1{-}p))$.
Unless otherwise stated, chess-match intervals are 95\% percentile intervals from 10,000 bootstrap resamples of opening clusters, keeping both colours together.
Table~\ref{tab:complement} row 2 clusters by single opening (1,000 groups), rows 1 and 4 by a shared 20-ply book prefix (556 groups), and rows 3 and 5 by an 8-ply prefix (411 groups).
Opponents in a comparison use the same openings, while scale selection has a reserved set and new comparisons draw fresh sets.
Appendix~\ref{app:reproduction} gives the full protocol.

\paragraph{Search-generated labels.}
A reference evaluator learns Stockfish 17.1's move preferences or position values in centipawns.
It tests what the same features and fitting procedure can capture from labels generated by a chess search.

\section{What the LLM Partnership Adds}

\subsection{Learning a useful search evaluator}
\label{sec:baselines}

Jev's labels contain information that a small evaluator can use in search.
On the mobility baseline, the Jev-trained evaluator adds 38.9 Elo \ci{29.9}{47.9} over 2,000 games.
Combining it with Qwen's labels gives a larger improvement: in direct 400-game matches, Jev+Qwen beats Jev alone by 77.7 Elo \ci{52.5}{103.7}, Qwen alone by 54.3 Elo \ci{32.2}{77.7}, and a mobility score rescaled to the same size by 84.1 Elo \ci{60.5}{109.5}.

The stronger baseline leaves less room for Jev alone.
The gains over $H_s$ are $+2.8$ \ci{-5.5}{11.0} for Jev, $+16.7$ \ci{9.0}{24.4} for Qwen, and $+33.8$ \ci{25.2}{42.4} for their combination.
The combined evaluator beats Qwen's in a direct 2,000-game match by 18.1 Elo \ci{10.8}{25.5} (640 wins, 824 draws, 536 losses).
A label source's value therefore depends on what the baseline and the other judge already supply.
Against Stockfish limited to Elo 2,400, the combined and published-table evaluators score $+376.9$ \ci{336.1}{424.3} and $+334.1$ \ci{296.3}{376.9}, respectively, in 400 games each on the same 200 openings.

\subsection{A partnership or more training pairs?}
\label{sec:budget}

A second judge competes with another use of the budget: more examples from the first judge.
We train Jev-only and Qwen-only evaluators on twice the pairs, giving each the same 252,540 responses and scale-selection budget as Jev+Qwen.
On 1,000 fresh openings played with both colours, the combination beats both (Table~\ref{tab:complement}).
Pooling two full-data matches against Qwen on twice the pairs gives $+9.6$ Elo \ci{4.3}{14.9} over 4,000 games; the individual estimates are $+8.0$ and $+11.1$.
A further 4,000-game match on fresh openings gives $+8.4$ \ci{2.9}{14.1} (1,152 wins, 1,793 draws, 1,055 losses).

\begin{table}[t]
\centering\small
\caption{Jev+Qwen vs.\ each rival at 252,540 responses, 2,000 games. Row 2 clusters by single opening; rows 1 and 4 by 20-ply prefix; rows 3 and 5 by 8-ply.}
\label{tab:complement}
\setlength{\tabcolsep}{3pt}\footnotesize
\begin{tabular}{@{}lrr@{}}
\toprule
Rival (252,540 responses each) & Elo & 95\% interval\\
\midrule
Jev, twice the pairs & $+27.0$ & \ci{19.2}{34.8}\\
Qwen, twice the pairs & $+8.0$ & \ci{0.9}{15.3}\\
Qwen, two answers averaged & $+14.3$ & \ci{5.9}{22.5}\\
Jev, two answers averaged & $+28.2$ & \ci{20.2}{36.3}\\
Qwen, adjusted toward 0.5 ($\tau{=}0.4465$) & $+8.0$ & \ci{0.5}{15.4}\\
\bottomrule
\end{tabular}
\end{table}

Repeated answers from one judge provide another equal-budget control.
Averaging two Jev answers, which agree on 92.1\% of pairs, or two Qwen answers still produces a weaker evaluator than Jev+Qwen.
Adjusting Qwen's probabilities to match the combination's average distance from 0.5 ($\tau{=}0.4465$) also leaves a gap (Table~\ref{tab:complement}).
In these comparisons, neither extra pairs nor repeated answers from one judge recover the benefit of the partnership.

\subsection{The information in Jev's labels}
\label{sec:substitutes}

Averaging probabilities changes both their values and their certainty.
We test whether those general changes explain the improvement by replacing Jev's contribution while keeping the response budget fixed (Table~\ref{tab:substitutes}; Figure~\ref{fig:headline}c).
Each comparison uses 4,000 games on fresh openings.

\begin{table}[t]
\centering\small
\caption{Jev+Qwen vs.\ four substitutes, 4,000 games on fresh openings at the same response budget.}
\label{tab:substitutes}
\setlength{\tabcolsep}{3pt}\footnotesize
\begin{tabular}{@{}lrr@{}}
\toprule
Substitute for Jev & Elo & 95\% interval\\
\midrule
Jev replaced by 0.5 & $+9.6$ & \ci{3.7}{15.6}\\
Jev probabilities shuffled freely & $+10.9$ & \ci{5.3}{16.8}\\
Qwen alone, recalibrated, twice the pairs & $+14.3$ & \ci{8.8}{20.2}\\
Jev shuffled within Qwen groups & $+5.3$ & \ci{-0.5}{11.0}\\
\bottomrule
\end{tabular}
\end{table}

Replacing Jev by 0.5 softens every Qwen label equally; freely shuffling Jev's probabilities keeps their distribution but breaks their connection to the position.
Both substitutions weaken the evaluator.
Recalibrating Qwen against Stockfish with a fitted temperature $T\approx18$ and using twice the pairs also trails the real combination.
These controls support a contribution from Jev's position-dependent judgments beyond uniform softening or this recalibration of Qwen.

A stricter shuffle moves Jev's probabilities only within ten equal-sized groups defined by Qwen's probability.
It preserves their association at that group level.
The combination's lead over this stricter substitute is $+5.3$ Elo \ci{-0.5}{11.0}.
The free shuffle shows that the match between Jev's probabilities and their positions carries the gain.

\subsection{Choosing a partner}
\label{sec:partners}

We next ask whether another model provides the same benefit alongside Qwen.
Each evaluator is fitted on three overlapping subsets, each containing two of three groups of source games and about half the pairs.
Within each run, all evaluators play the corresponding Qwen evaluator on the same 1,000 colour-paired openings.
We compare two rules for $\lambda$: retain the full-data setting (300 for the combinations, 100 for Qwen on twice the pairs), or select it anew within each subset.

\begin{table}[t]
\centering\small
\caption{Jev+Qwen's lead over alternative Qwen partnerships, 6,000 games per evaluator across three subsets. Intervals resample openings.}
\label{tab:partners}
\setlength{\tabcolsep}{3pt}\footnotesize
\begin{tabular}{@{}lrr@{}}
\toprule
Jev+Qwen's lead over & Elo & 95\% interval\\
\midrule
\multicolumn{3}{@{}l}{\textit{Full-data settings}}\\
OLMo+Qwen & $+16.6$ & \ci{10.3}{22.8}\\
DiffusionGemma+Qwen & $+16.9$ & \ci{10.2}{23.4}\\
\midrule
\multicolumn{3}{@{}l}{\textit{Settings selected per subset}}\\
OLMo+Qwen & $+7.7$ & \ci{1.4}{14.0}\\
DiffusionGemma+Qwen & $+7.9$ & \ci{1.3}{14.6}\\
\bottomrule
\end{tabular}
\end{table}

\begin{figure}[t]
\centering
\includegraphics[width=\columnwidth]{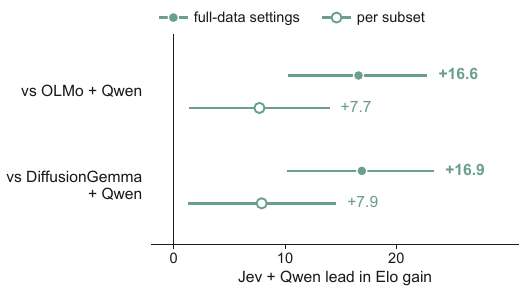}
\caption{Jev+Qwen's lead over alternative partnerships, 6,000 games per evaluator. Filled: full-data settings; open: per-subset. Lines: 95\% intervals.}
\label{fig:partners}
\end{figure}

Jev+Qwen leads OLMo+Qwen and DiffusionGemma+Qwen under both rules (Table~\ref{tab:partners}).
The lead is smaller when each subset selects its own penalty.
Under that rule, Jev+Qwen's advantage over Qwen on twice the pairs is $+2.2$ Elo \ci{-2.4}{6.8}.
The full-data result is $+9.6$ Elo \ci{4.3}{14.9} (Section~\ref{sec:budget}).
Figure~\ref{fig:partners} and Appendices~\ref{app:penalty}--\ref{app:per_replicate} show the effect of these training choices.
Testing all combinations of the four judges likewise gives no simple benefit from adding more judges (Appendix~\ref{app:panel}).

\subsection{Why measure decisions after search?}
\label{sec:mechanism}

A label-level score does not explain the ordering of the resulting players.
On 12,280 held-out pairs with a non-tied Stockfish preference, agreement is similar across judges: 54.5\% for Jev, 52.9\% for Qwen, 53.4\% for OLMo, and 54.4\% for DiffusionGemma.
Jev has the lowest expected calibration error against those preferences, 0.081, compared with 0.159 for DiffusionGemma, 0.176 for OLMo, and 0.260 for Qwen.

Jev and Qwen disagree on 30.9\% of the pairs (3,791 comparisons).
On those disagreements, Jev matches Stockfish 53.6\% of the time (95\% interval \ci{51.5}{55.6}, resampling source games).
The averaged probabilities match only 48.2\%, yet they train the stronger full-data player.
The same mismatch appears after fitting: on 1,285 quiet positions, Jev's evaluator has higher move accuracy than the combined evaluator, 62.4\% versus 60.7\%, while losing in direct play.
Appendix~\ref{app:static} gives the paired tests and a similar comparison using evaluators trained on Stockfish.

The evaluator supplies a score repeatedly within a search tree, where the final action depends on many evaluated continuations.
The playing-strength test measures what static accuracy does not: how the score performs across a full search tree.

\section{Beyond Chess: Passage Reranking}
\label{sec:retrieval}

The labels idea extends beyond move evaluation.
For text search, the choices are passages for a query, the reranker is a MiniLM reranker~\cite{wang2020minilm}, and success is agreement with human relevance judgments.
This provides a second test of what can be built from probability-only answers and whether an LLM partner improves it.

On 500 human-graded TREC Deep Learning 2019 passage pairs~\cite{craswell2020trecdl,craswell2021trecdl}, Jev chooses the more relevant passage 86.2\% of the time and Qwen 87.2\%, compared with 62.1\% for BM25.
We then collect both judges' probabilities on 80,000 MS~MARCO pairs~\cite{bajaj2018msmarco}, in both presentation orders, and train rerankers for 10,000 steps with five seeds.
Each reranker reranks BM25's top 100 passages for 97 human-judged TREC Deep Learning 2019 and 2020 queries.
It ranks independently after training, with no judge calls.

\begin{figure}[t]
\centering
\includegraphics[width=\columnwidth]{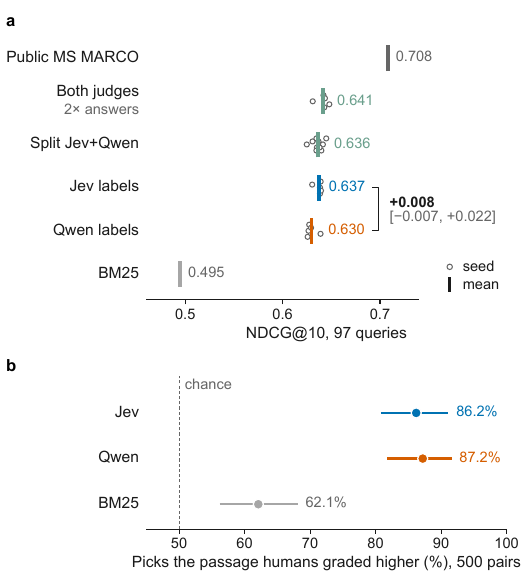}
\caption{(a)~Passage-reranking NDCG@10 on TREC DL 2019+2020. Open dots: individual training runs; tick: mean over five seeds, with both order assignments pooled for the split row. (b)~Agreement with human graders on 500 pairs.}
\label{fig:retrieval}
\end{figure}

\paragraph{Useful labels from Jev alone.}
Jev's reranker scores 0.637 in NDCG@10, a ranking-quality metric where higher is better; Qwen's scores 0.630 (unrounded, 0.6371 and 0.6295).
Their difference is $+0.008$ \ci{-0.007}{0.022} with seeds and queries resampled: the two judges finish within 0.008 of each other.
Jev's reranker improves on BM25's 0.495 by $+0.142$ \ci{0.098}{0.186}.
A public MS~MARCO cross-encoder of the same size scores 0.708.
Collecting Jev's 160,000 answers took about 21 minutes of API time with no local GPU; Qwen required about 5.1 GPU-hours in our labeling setup.

\paragraph{Spending the budget on coverage or another judge.}
Adding Qwen to Jev gains at most $+0.004$ \ci{-0.003}{0.011} in any tested combination.
Splitting the two presentation orders between Jev and Qwen gives similar quality at the same answer budget.
Using both judges on every pair doubles the responses and adds $+0.004$ \ci{-0.003}{0.011}, with queries resampled.
When the budget instead forces a choice between two judges on half the pairs and Jev on all the pairs, broader coverage wins: the combination loses $0.032$ in NDCG@10 (95\% interval \ci{0.018}{0.045}) under matched 5,000-step training.
Asking Qwen only about Jev's least certain pairs loses to Jev alone at 80,000 answers and does not beat it at 160,000.
Appendix~\ref{app:retrieval} reports these allocations and their intervals.

Chess and retrieval support the same approach but lead to different judge choices.
The chess evaluator gains from the partnership at full data; the reranker trains as well from Jev alone, and adding Qwen gains at most $+0.004$.

\section{What the Proof of Concept Shows}
\label{sec:discussion}

\paragraph{Reasoning through the surrounding system.}
The chess system obtains lookahead from search and positional guidance from learned judgments.
Distillation makes that guidance cheap enough to reuse throughout the tree, and the deployed bot also draws on opening and endgame tools.
The reasoning is the system's explicit search over continuations, guided by the learned evaluator.
This separation of roles makes a probability-only interface useful even when the task requires considering consequences beyond the immediate choice.

\paragraph{Useful judgment is task dependent.}
A search can repeatedly use an unhelpful score as readily as a useful one.
Jev-derived evaluation improves play in chess and Breakthrough but weakens it in Othello and Connect-4 (Appendix~\ref{app:games}).
The Othello checks expose a concrete problem: the judges rarely prefer corner moves in the tested comparisons, and the Jev-trained evaluator assigns corners a negative weight (Appendix~\ref{app:othello}).
The labels need to contain knowledge suited to the task.

\paragraph{A partnership is a choice about labels.}
The chess controls show that Jev's contribution survives comparisons with repeated answers, extra pairs, and several substitutes.
In chess, combining Jev and Qwen produces a stronger player; in passage reranking, Jev alone scores as high as Qwen and adding Qwen gains at most $+0.004$.
Spend responses on another judge only when its labels improve the evaluator enough to justify the cost.
Pairwise probabilities can support both a small search evaluator and a text reranker, with the model chosen for the decisions it will make.

\section{Labeling Cost and Reproduction}
\label{sec:cost}

Moving judgment offline separates the cost of collecting labels from the cost of making a decision.
Table~\ref{tab:cost} reports the measured configurations for 126,270 chess responses, covering 63,135 pairs in both orders.
Jev cost \$1.97 and about 16 minutes of API time at TypeSafe's stated \$0.042 per million input tokens, with output free, 372.3 billed tokens per call, and no local GPU.
Qwen and OLMo each used over an hour of GPU time while answering sequentially.
DiffusionGemma, served through djev on one RTX~5090, completed the labels in 0.28 GPU-hours.
Jev's row is API wall-clock time; the others are GPU time on the listed hardware.
After fitting, all chess evaluators use the same 443-weight form during play.
Appendices~\ref{app:fit}, \ref{app:reproduction}, \ref{app:prompt}, and \ref{app:retrieval} specify the splits, features, prompts, and training procedures.

\begin{table}[t]
\centering\small
\caption{Cost of obtaining 126,270 judge responses (63,135 pairs, both orders). Effective rates reflect concurrent requests.}
\label{tab:cost}
\setlength{\tabcolsep}{3pt}\footnotesize
\begin{tabular}{@{}llrlr@{}}
\toprule
Judge & ms/resp & Timing & Hardware & Cost\\
\midrule
Jev & 7.6 & 20 concurrent & API; no local GPU & \$1.97\\
Qwen3-32B & 54.6 & sequential & 1$\times$ H200 & 1.91 GPU-h\\
OLMo-2-32B & 35.6 & sequential & PRO 6000 + H200 & 1.25 GPU-h\\
DiffusionGemma & 7.9 & 32 concurrent & 1$\times$ RTX 5090 & 0.28 GPU-h\\
\bottomrule
\end{tabular}
\end{table}

\section{Related Work}

\paragraph{Distillation and learned feedback.}
Hinton et al.\ train a smaller network to match averaged output probabilities~\cite{hinton2015distilling}, and You et al.\ study learning from multiple judges~\cite{you2017multiteacher}.
Motif uses collected language-model preferences to learn a reward for reinforcement learning~\cite{klissarov2023motif}.
We use probability averaging and preference fitting as established tools, and study what a probability-only judge contributes when its evaluator is used in a decision system.
The comparisons hold model size, search, and response budget fixed.

\paragraph{Combining imperfect judgments.}
Dawid and Skene model annotator error rates~\cite{dawid1979}, while Snorkel combines imperfect label sources~\cite{ratner2017snorkel}.
WARM averages reward-model weights to reduce exploitation of the learned reward~\cite{rame2024warm}.
Our evaluators learn from a simple average of probabilities, allowing the experiments to focus on the value of the additional judge and its labels.

\paragraph{LLMs and probability-only models as judges.}
Zheng et al.\ evaluate language models as judges of human preferences~\cite{zheng2023judging}; Verga et al.\ combine diverse smaller judges in panels~\cite{verga2024juries}.
Rao and Callison-Burch compare Jev and LLMs on rubric judgments and identify correlated failures~\cite{rao2026jevvsllms}.
Li et al.\ accept confident Jev judgments and send uncertain cases to a reasoning judge at query time~\cite{li2026jevjudge}.
Our setting transfers judgments into an evaluator that runs without either judge.
The retrieval gate tests an offline version of uncertainty-based allocation.

\paragraph{Evaluation and planning in games.}
TDLeaf($\lambda$) learns evaluation weights from search leaves~\cite{baxter1999tdleaf}, and Texel tuning fits them to game outcomes~\cite{osterlund2014texel}.
DeepChess learns pairwise comparisons for alpha-beta search~\cite{david2016deepchess}; SentiMATE learns move evaluations from human commentary~\cite{kamlish2019sentimate}.
NNUE embeds an efficiently updatable neural evaluator in traditional search~\cite{nasu2018nnue}.
Ruoss et al.\ train a transformer on Stockfish scores to play without search~\cite{ruoss2024amortized}, while Schultz et al.\ study external and internal planning with language models in board games~\cite{schultz2024mav}.
Our proof of concept distills fixed-choice model judgments into a compact evaluator and lets search supply the lookahead.

\section{Conclusion}

Probability-only judgments are useful components of systems that plan and act under time constraints.
The chess proof of concept connects Jev-derived evaluation to search and supporting tools, then distills judgments into weights that search can reuse without live model calls.
Combining Jev and Qwen improves the full-data evaluator even when either judge alone receives the same response budget.
In passage reranking, Jev alone trains a useful reranker, and adding Qwen gains at most $+0.004$ in NDCG@10.
Each component has a concrete role: search considers consequences, distillation makes judgment reusable, and an additional judge earns its budget by what it contributes to the deployed evaluator.

\small
\bibliographystyle{unsrtnat}
\bibliography{refs}
\normalsize
\appendix

\section{Fitting Details}
\label{app:fit}

We hold out one fifth of source games for evaluation (seed 8) and, within the rest, one fifth for selecting training settings (seed 9).
Each feature is divided by its standard deviation in the training data; constant features are removed.
The candidate values for the weight-size penalty $\lambda$ are $\{1,10,100,300,1000,3000,10000,30000\}$.
Newton updates stop after 30 iterations or a maximum weight change below $10^{-8}$.
Exported weights are centred within each piece-square block, scaled to a standard deviation of one on 1,000 fixed positions, and rounded to integers with denominator 4096.
The reference evaluator trained on Stockfish scores uses targets $\sigma(kv)$, where $v$ is the position score in centipawns and $k=0.00368208$.
Material contributes a fixed offset in log odds.

\paragraph{Single-judge evaluators with the same response budget.}
The second training block contains the first 63,135 pairs from a fixed ordering of newly labelled positions.
Jev and Qwen receive the same board format and question wording.
The two blocks together draw on 2,145 source games, against 2,042 for the first block alone, so the second block adds new positions from mostly the same source games.
Jev answered the second block with all pairs of a position in one request; Qwen answered each item separately, as in the first block.
Each single-judge evaluator trains on its labels from both blocks, totalling 126,270 pairs, using the same code as the main evaluators.
Running this code on the original single-judge training sets reproduces their weight files exactly.
The selected weight-size penalties are 10 for Jev and 100 for Qwen.

\section{Scale Selection}
\label{app:sweep}

Table~\ref{tab:sweep} lists every validation setting against the unchanged published tables, 200 games each on the same 100 colour-paired openings.
Every later comparison uses the scale marked in bold.

Table~\ref{tab:robust} tests the single-judge evaluators trained on twice the pairs at alternative scales, using 200 unused openings.

\begin{table}[h]
\centering\small
\caption{Jev+Qwen against single-judge evaluators trained on twice the pairs at the listed scales, 400 games on 200 unused openings.}
\label{tab:robust}
\begin{tabular}{@{}lrr@{}}
\toprule
Evaluator and scale $c$ & Elo & 95\% interval\\
\midrule
Jev, $c=0.5$ & $+32.2$ & \ci{13.9}{50.7}\\
Jev, $c=0.125$ & $+27.9$ & \ci{10.4}{46.3}\\
Qwen, $c=0.5$ & $+19.1$ & \ci{4.3}{34.0}\\
Qwen, $c=0.125$ & $+34.9$ & \ci{16.5}{53.4}\\
\bottomrule
\end{tabular}
\end{table}

\begin{table}[h]
\centering\small
\caption{Validation Elo against published tables. Each coefficient is the ratio of the learned evaluator's standard deviation to the baseline's. Bold marks the selected scale.}
\label{tab:sweep}
\setlength{\tabcolsep}{4pt}
\begin{tabular}{@{}rrrrrr@{}}
\toprule
 & & \multicolumn{2}{c}{Same pairs} & \multicolumn{2}{c}{Twice the pairs}\\
\cmidrule(lr){3-4}\cmidrule(l){5-6}
$c$ & Jev+Qwen & Jev & Qwen & Jev & Qwen\\
\midrule
0.125 & $+33.1$ & $+10.4$ & $\mathbf{+24.4}$ & $+10.4$ & $+29.6$\\
0.25 & $+31.4$ & $+10.4$ & $0.0$ & $\mathbf{+24.4}$ & $\mathbf{+34.9}$\\
0.5 & $\mathbf{+57.9}$ & $\mathbf{+12.2}$ & $+17.4$ & $-3.5$ & $+17.4$\\
1.0 & $-5.2$ & $-24.4$ & $-125.0$ & $-24.4$ & $-68.6$\\
\bottomrule
\end{tabular}
\end{table}

\section{The Recorded Lichess Deployment}
\label{app:live}

Figure~\ref{fig:lichess} records a segment of the bot's deployment against other Lichess bots at one minute per side with no increment.
The segment contains 74 wins, 10 draws, and 16 losses, with a rating rise from 1753 to 2268.
The deployed bot combines Stockfish search, material count, a 411-weight Jev-trained evaluator, Syzygy endgame tablebases~\cite{syzygy}, a masters opening book, and live Jev choices among near-tied moves.
The opening book supplies established continuations, tablebases supply exact information for covered endgames, and the evaluator guides search outside those resources.
The controlled study uses the separate 443-weight evaluator and disables the book, tablebases, and live queries during play.

\section{Testing All Judge Combinations}
\label{app:panel}

To test whether more judges give a better evaluator, Table~\ref{tab:panel} covers all 15 nonempty combinations of the four judges.
Each evaluator trains on the average of its judges' probabilities for 63,135 pairs, uses the same scale-selection procedure and plays 1,000 games against the published tables on the same 500 openings.

\begin{table}[h]
\centering\small
\caption{All 15 judge combinations, 1,000 games each against the published-table evaluator on shared openings. $n$: number of judges.}
\label{tab:panel}
\setlength{\tabcolsep}{3pt}\footnotesize
\begin{tabular}{@{}llrr@{}}
\toprule
Judges & $n$ & Elo & 95\% interval\\
\midrule
Jev+Qwen & 2 & $+25.8$ & \ci{12.9}{38.4}\\
Qwen & 1 & $+24.0$ & \ci{13.2}{34.9}\\
Jev+Qwen+DiffusionGemma & 3 & $+19.1$ & \ci{8.0}{30.3}\\
Qwen+OLMo+DiffusionGemma & 3 & $+13.6$ & \ci{2.1}{25.1}\\
Jev & 1 & $+13.2$ & \ci{2.1}{24.0}\\
OLMo+Qwen & 2 & $+13.2$ & \ci{2.4}{24.7}\\
All four & 4 & $+13.2$ & \ci{2.1}{24.4}\\
DiffusionGemma+Qwen & 2 & $+11.1$ & \ci{0.7}{21.9}\\
DiffusionGemma & 1 & $+10.4$ & \ci{-0.3}{21.6}\\
Jev+Qwen+OLMo & 3 & $+6.3$ & \ci{-4.5}{17.4}\\
OLMo & 1 & $+5.9$ & \ci{-4.5}{16.3}\\
OLMo+DiffusionGemma & 2 & $+4.5$ & \ci{-6.6}{15.6}\\
Jev+DiffusionGemma & 2 & $-2.8$ & \ci{-13.2}{7.6}\\
Jev+OLMo+DiffusionGemma & 3 & $-3.5$ & \ci{-14.6}{8.0}\\
Jev+OLMo & 2 & $-5.9$ & \ci{-16.7}{4.9}\\
\bottomrule
\end{tabular}
\end{table}

\section{Locating the Learned Contribution}
\label{app:weight_swap}

We swap groups of weights between the full-data Jev+Qwen and DiffusionGemma+Qwen evaluators to test whether one feature family explains the difference between these partnerships.
Each modified evaluator replaces one group of weights, covering piece-square values, mobility, pawn structure, king safety, threats or other features, while retaining the original evaluator's scale.
Swapping piece-square weights changes the learned evaluator's standard deviation to $0.33\times$ the original in one direction and $3.9\times$ in the other.
Swapping any other group leaves it between $0.98\times$ and $1.13\times$ the original.
Each modified evaluator plays 1,000 games against the published tables on a shared set of 500 openings (Table~\ref{tab:weight_swap}).

We also rescale the substituted piece-square weights to keep the learned evaluator's total standard deviation equal to the original evaluator's.
With total score variation held fixed, swapping piece-square weights changes Jev+Qwen by $-8.7$ \ci{-25.8}{8.4} and DiffusionGemma+Qwen by $+9.7$ \ci{-5.2}{24.7}.

\begin{table}[h]
\centering\small
\caption{Elo change from replacing one weight group, 1,000 games per row. Piece-square swaps change total score variation; rescaled swaps hold it fixed.}
\label{tab:weight_swap}
\setlength{\tabcolsep}{3pt}\footnotesize
\begin{tabular}{@{}lrr@{}}
\toprule
Weight group & $\Delta$Elo & 95\% interval\\
\midrule
\multicolumn{3}{@{}l}{\textit{Jev+Qwen, replacing from DiffusionGemma+Qwen}}\\
\midrule
Piece-sq.\ ($0.33\times$ scale) & $-7.7$ & \ci{-23.8}{8.4}\\
Piece-sq.\ (matched scale) & $-8.7$ & \ci{-25.8}{8.4}\\
mobility & $+3.1$ & \ci{-13.6}{19.9}\\
king safety & $-10.1$ & \ci{-27.2}{6.6}\\
pawn structure & $-7.3$ & \ci{-24.1}{9.7}\\
threats & $-10.5$ & \ci{-28.6}{8.0}\\
other & $-4.5$ & \ci{-21.3}{12.6}\\
\midrule
\multicolumn{3}{@{}l}{\textit{DiffusionGemma+Qwen, replacing from Jev+Qwen}}\\
\midrule
Piece-sq.\ ($3.9\times$ scale) & $-4.5$ & \ci{-19.8}{11.1}\\
Piece-sq.\ (matched scale) & $+9.7$ & \ci{-5.2}{24.7}\\
mobility & $-0.7$ & \ci{-15.6}{14.3}\\
king safety & $+10.4$ & \ci{-4.5}{25.4}\\
pawn structure & $+1.0$ & \ci{-14.3}{16.4}\\
threats & $+13.9$ & \ci{-0.3}{28.9}\\
other & $+9.0$ & \ci{-5.9}{24.4}\\
\bottomrule
\end{tabular}
\end{table}

\section{Results at Other Time Controls}
\label{app:clock}

\begin{figure*}[t]
\centering
\includegraphics[width=\textwidth]{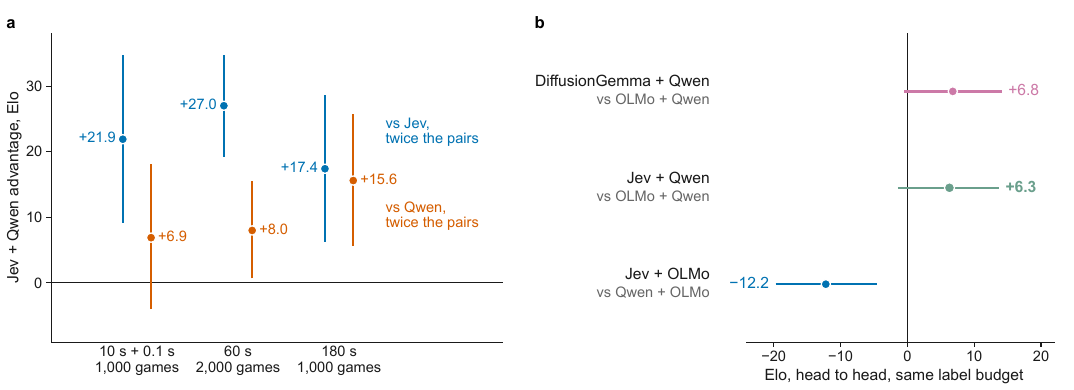}
\caption{(a)~Jev+Qwen vs.\ single-judge evaluators on twice the pairs at three time controls. (b)~Changing one judge at equal budget, 2,000 games each. Dots: estimates; lines: 95\% intervals.}
\label{fig:robustness}
\end{figure*}

At 180 seconds per side, Jev+Qwen leads Qwen trained on twice the pairs by 15.6 Elo \ci{5.7}{25.8} and Jev trained on twice the pairs by 17.4 Elo \ci{6.3}{28.7}.
At 10 seconds per side plus 0.1 seconds per move, the Elo differences are $+6.9$ \ci{-4.0}{18.1} against Qwen and $+21.9$ \ci{9.2}{34.7} against Jev.
Figure~\ref{fig:robustness}b changes one judge at a time in an earlier single-subset comparison: Jev+Qwen leads OLMo+Qwen by $+6.3$ \ci{-1.4}{13.7}, and Jev+OLMo trails Qwen+OLMo by $-12.2$ \ci{-19.7}{-4.5}; Jev's partner advantage is measured with Qwen.

\section{Transfer to Other Games}
\label{app:games}

\begin{figure}[h]
\centering
\includegraphics[width=\columnwidth]{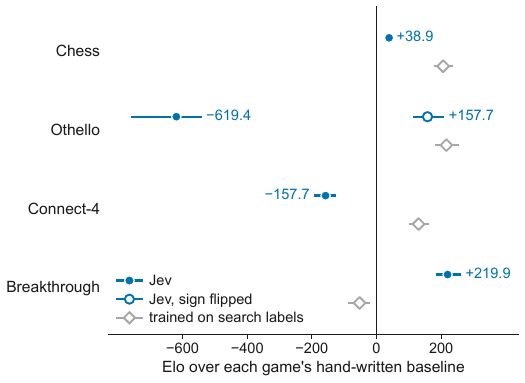}
\caption{Playing-strength change from adding a Jev-trained score. Filled: 400-game matches (chess: 2,000; open: sign-flipped Othello, 200). Diamonds: engine-search labels. Lines: 95\% intervals.}
\label{fig:knowledge}
\end{figure}

The value of distilled judgment depends on the game.
Adding a Jev-trained score improves chess and Breakthrough and weakens Othello and Connect-4.
We apply the same comparison in three additional games: each game's hand-written evaluator, that evaluator plus a Jev-trained score, and a reference score trained from search-generated labels.
Jev is the only judge whose labels train these scores (Figure~\ref{fig:knowledge}).
In chess, adding the Jev-trained score improves the mobility baseline by 38.9 Elo \ci{29.9}{47.9} over 2,000 games.
Training the same features on Stockfish preferences adds 206.2 Elo.
Jev-trained scores reduce strength in Othello and Connect-4; the Connect-4 change is $-157.7$ Elo.
Reference scores using the same features add $+216.1$ and $+129.9$ Elo, respectively.
In Othello, subtracting the Jev-trained score improves play.
On 5,856 corner-versus-non-corner pairs, Jev prefers the corner 14.4\% of the time.
All three tested judges prefer the non-corner move in most pairs (Qwen 17.7\%, OLMo 15.3\%; Jev on a coordinate-list rendering 24.3\%; Appendix~\ref{app:othello}).
Jev's low preference for corner moves accompanies a loss of 619.4 Elo over 400 games when its trained score is added to the baseline.
In Breakthrough, the Elo changes are $+219.9$ for the Jev-trained score and $-52.5$ for the reference score.
Jev's gain is $+25.1$ above the random scores' 95th-percentile gain of $+194.8$; the random scores were tested on 50 of the 200 openings.

\section{Move Accuracy on Fixed Positions}
\label{app:static}

On 1,285 quiet positions where Stockfish separates the two leading candidates by at least 80 centipawns, the Jev evaluator selects the better move 62.4\% of the time and Jev+Qwen 60.7\%.
Jev has the higher move accuracy (exact McNemar $p{=}0.15$), while Jev+Qwen plays more strongly in direct games.
Including material and mobility, move accuracy is 65.29\% for Jev and 64.44\% for Jev+Qwen.
The Jev evaluator trained on twice the pairs selects 808 moves correctly against Jev+Qwen's 780 ($p{=}0.07$), while Jev+Qwen wins their direct match by 27.0 Elo.

Two Stockfish-trained evaluators using the same features also rank differently on move accuracy and playing strength (Table~\ref{tab:static}).
On 12,280 held-out move comparisons, the evaluator trained on Stockfish position scores answers 58.1\% correctly, compared with 61.4\% for the evaluator trained on Stockfish move preferences.
The test uses move comparisons, the same task the preference evaluator trains on.
Against the same mobility baseline, the evaluator trained on position scores gains 254.7 Elo and the evaluator trained on move preferences 206.2.
The difference between their gains is 48.5 Elo \ci{7.9}{88.4}, measured on shared openings.

\begin{table}[h]
\centering\small
\caption{Move accuracy (\%) and playing strength. The complete evaluator includes material, mobility and the integer calculations used during play.}
\label{tab:static}
\begin{tabular}{@{}lrr@{}}
\toprule
1,285 quiet positions & Jev & Jev+Qwen\\
\midrule
Learned evaluator & 62.4 & 60.7\\
Complete evaluator & 65.29 & 64.44\\
Jev+Qwen lead, Elo & & $+77.7$\\
\midrule
12,280 Stockfish-labelled pairs & Preferences & Scores\\
\midrule
Learned evaluator & 61.41 & 58.13\\
Complete evaluator & 59.63 & 58.79\\
Gain over mobility, Elo & $+206.2$ & $+254.7$\\
\bottomrule
\end{tabular}
\end{table}

\section{Choosing the Weight-Size Setting}
\label{app:penalty}

Several weight-size settings reach nearly the same validation loss, so the partner study reports two rules: each evaluator keeps its full-data setting, or each training subset selects its own.
Jev+Qwen leads OLMo+Qwen and DiffusionGemma+Qwen under both (Table~\ref{tab:partners}).
With settings selected per subset on half the data, Jev+Qwen against Qwen trained on twice the pairs is $+2.2$ \ci{-2.4}{6.8}; the full-data matches (Section~\ref{sec:budget}) measure this comparison at $+9.6$ \ci{4.3}{14.9}.

We recalibrate each partner's probabilities against Stockfish, then average them with Qwen's original probabilities.
The Elo changes are $+3.8$ \ci{-2.1}{9.6} for OLMo+Qwen, $+6.1$ \ci{-0.2}{12.4} for DiffusionGemma+Qwen and $-2.1$ \ci{-8.6}{4.4} for Jev+Qwen.

\section{Results for Each Training Subset}
\label{app:per_replicate}

Table~\ref{tab:per_split} gives each training subset's results using the fixed weight-size settings.

\begin{table}[h]
\centering\small
\caption{Elo for each evaluator against the corresponding Qwen evaluator trained on 2$\times$ the pairs, with 2,000 games per comparison. Each evaluator retains its full-data weight-size setting.}
\label{tab:per_split}
\setlength{\tabcolsep}{3pt}\footnotesize
\begin{tabular}{@{}lrrr@{}}
\toprule
Evaluator & Subset 1 & Subset 2 & Subset 3\\
\midrule
Jev+Qwen & $+14.8$ & $+7.8$ & $+9.4$\\
OLMo+Qwen & $-10.3$ & $-4.7$ & $-2.8$\\
DiffusionGemma+Qwen & $-12.7$ & $-2.3$ & $-3.8$\\
Qwen + Jev shuffled within Qwen groups & $-1.6$ & $+3.8$ & $+4.2$\\
\bottomrule
\end{tabular}
\end{table}

\section{Examining the Othello Failure}
\label{app:othello}

The Othello checks examine whether the poor result arises from invalid choices or from the judgments themselves.
All 127,154 candidate moves in the Othello labels are legal for the side to move in the supplied board text.
In a hand-checked position with White to move, the corner a8 captures four discs along the long diagonal; Jev assigns it probability 0.27 against the non-corner move e6.
Across 5,856 pairs comparing a corner move with a non-corner move, Jev prefers the corner in 14.4\% and assigns it a mean probability of 0.42.
The deeper reference search labels 2,950 of these pairs and assigns the corner a mean probability of 0.69.
Among the three judges tested, corner-preference rates are 17.7\% for Qwen3-32B and 15.3\% for OLMo; Jev's rate is 24.3\% when the board is presented as a coordinate list.
Training on Jev's answers produces a corner weight of $-838$; the baseline assigns $+858$.
On 100 colour-paired openings, the baseline plus Jev's evaluator scores 5 wins, 2 draws and 193 losses ($-603.9$ Elo in this separate 200-game check, bootstrap interval \ci{-798.3}{-502.6}); the baseline minus that evaluator scores 141 wins, 3 draws and 56 losses ($+157.7$ Elo, \ci{111.4}{209.9}).
Doubling the baseline's corner weight gains $+54.3$ \ci{13.9}{96.2}.
Subtracting the full Jev-trained evaluator, including its other nine weights, gives a larger measured gain.

\section{Reproduction Details}
\label{app:reproduction}

\paragraph{Candidate moves.}
Candidates come from a material-only Stockfish 17.1 search (piece values 100, 300, 300, 500 and 900 centipawns, correction history off) run with MultiPV~4, 0.1\,s per position and one thread.
A candidate is kept when it scores within 20 centipawns of the best line; positions whose best line is a mate are dropped.
The median search depth across all searched positions is 17.
On the 13,000 selected positions, 72.4\% have the maximum of four candidates; the mean is 3.57.

\paragraph{Features.}
Table~\ref{tab:features} groups the 443 features.

\begin{table}[h]
\centering\small
\caption{The 443 features by family.}
\label{tab:features}
\setlength{\tabcolsep}{4pt}\footnotesize
\begin{tabular}{@{}lrl@{}}
\toprule
Family & Count & Examples\\
\midrule
Piece-square & 384 & 6 piece types $\times$ 64 squares\\
Mobility & 8 & attacked squares, safe mobility\\
King safety & 18 & zone attacks, shield pawns, castled\\
Pawn structure & 16 & doubled, isolated, passed, backward\\
Threats & 11 & hanging pieces, attacked by lower value\\
Other & 6 & bishop pair, rook on open file, outposts\\
\bottomrule
\end{tabular}
\end{table}

\paragraph{Source games.}
The first block's 13,000 positions come from 2,042 games: 1,835 from the Lichess monthly database (August 2026, rated bullet and blitz, both players rated at least 2200) and 207 from the lab's own bot accounts on Lichess.
Positions are sampled every fourth ply from ply 12 to 120, excluding positions in check and those with fewer than two legal moves.
Test-bank positions are excluded before sampling.
Of the 30,679 extracted positions, 21,104 have at least two candidates; 13,000 of these are selected by a fixed shuffle (seed~8).

\paragraph{Opening suites.}
Each comparison uses fresh openings drawn from a 20-ply masters opening book (Lichess opening explorer, masters and rated players above 2000).
The 22-ply sets extend book leaves by two uniformly random legal plies.
The 12-ply sets draw one of the book's 8-ply lines and extend it by four random plies, keeping only positions with equal material.
Each generator rejects any position or move sequence already used in an earlier set; no two sets share a final position.

\paragraph{Bootstrap intervals.}
Unless otherwise stated, chess-match intervals are 95\% percentile intervals from 10,000 bootstrap resamples of opening clusters (seed 20260928).
The pooled headline result, the third replicate and the substitute tests cluster by single opening.
Table~\ref{tab:complement} row 2 clusters by single opening (1,000 groups); rows 1 and 4 by 20-ply book prefix (556 groups); rows 3 and 5 by 8-ply prefix (411 groups).

\section{Prompt}
\label{app:prompt}

Each query opens with ``Position to judge:'', then lists the occupied squares of each side in piece order (e.g.\ \texttt{White: Ke1 Qd1 Ra1 Rh1 \ldots{} e4 d4}), the side to move, and the two moves being compared in standard algebraic notation (e.g.\ \texttt{Candidate moves for White: Nxf6, Bd3}).
The question offers the two listed moves as answer choices: ``Which of these two moves is better for [White/Black]?''
Both presentation orders are queried and the probabilities averaged.
Jev and DiffusionGemma receive the question through the Choice interface, which requests probabilities for the specified answers.
Qwen and OLMo receive a chat prompt whose allowed answers are the two moves.
For the second set of labels, each Jev request presents all candidate moves for one position.

\section{Retrieval Details and Comparisons}
\label{app:retrieval}

We sample 80,000 passage pairs from 8,000 MS~MARCO training queries, 10 pairs per query, and label each pair in both presentation orders with Jev and Qwen3-32B.
Qwen receives the same passage prompt as Jev and a one-line instruction to answer A or B; its label is the probability it assigns to each answer.
Each reranker initializes from MiniLM-L6-H384-uncased and trains for 10,000 steps in the main comparison and 5,000 steps in the comparisons that trade pairs for Qwen's answers (batch size 32, learning rate $2\times10^{-5}$) on Bradley--Terry soft labels, with 5 random seeds.
The gate selects pairs by Jev's certainty $|p-0.5|$ averaged over both orders; a pair whose two orders disagree counts as least certain.
At 5,000 steps and 80,000 answers (40,000 pairs), Jev scores 0.550 and Qwen 0.560.
Doubling the training steps from 5,000 to 10,000 at 160,000 answers adds $+0.038$ for Jev and $+0.030$ for Qwen.

\paragraph{Order consistency.}
On the 500 human-graded passage pairs, Jev selects the same answer in both orders on 94.8\% of pairs and Qwen on 90.6\%.
On the 80,000 training pairs, the corresponding rates are 86.7\% and 72.0\%.

\paragraph{Combining judges at fixed pair coverage.}
Splitting the two presentation orders between the judges uses 80,000 answers from each and scores 0.636 in NDCG@10, a difference of $-0.001$ \ci{-0.011}{0.009} from Jev alone.
Using both judges on every pair uses 320,000 answers and scores 0.641, a difference of $+0.004$ \ci{-0.003}{0.011}, with queries resampled.
These comparisons use 10,000 training steps.

\paragraph{Trading coverage for additional judgments.}
With 5,000 steps and 160,000 total answers, averaging both judges on 40,000 pairs loses $0.032$ \ci{0.018}{0.045} to Jev alone on 80,000 pairs, with queries resampled.
The gate asks Qwen about the 25\% of pairs on which Jev is least certain.
It loses 0.047 to Jev alone at 80,000 answers (\ci{0.013}{0.084}) and gives a difference of $-0.007$ \ci{-0.027}{0.012} at 160,000 answers, with seeds and queries resampled.
At 80,000 answers, dropping 20\% of pairs costs 0.050, while Qwen's labels on uncertain pairs recover only 0.004.

\end{document}